%% file: ms.tex
\documentclass[10pt,twocolumn,letterpaper]{article}
\usepackage{pdfpages}
\usepackage{iccv}
\usepackage{times}
\usepackage{epsfig}
\usepackage{graphicx}
\usepackage{amsmath,bm}
\usepackage{amssymb}
\usepackage{array}
\newcommand{\argmax}{\mathop{\mathrm{argmax}}}
\usepackage[pagebackref=true,breaklinks=true,letterpaper=true,colorlinks,bookmarks=false]{hyperref}

\iccvfinalcopy % *** Uncomment this line for the final submission

\def\iccvPaperID{9206} % *** Enter the ICCV Paper ID here

\ificcvfinal\fi

\begin{document}

%%%%%%%%% TITLE
\title{Memory-Augmented Non-Local Attention for Video Super-Resolution}

\author{Jiyang Yu\textsuperscript{1},\ \ \ \ Jingen Liu\textsuperscript{1},\ \ \ \  Liefeng Bo\textsuperscript{2},\ \ \ \  Tao Mei\textsuperscript{3}\\
\textsuperscript{1}JD AI Research, Mountain View, USA,\\ \textsuperscript{2}JD Finance America Corporation, Mountain View, USA\\
\textsuperscript{3}JD AI Research, Beijing, China\\
{\tt\small \{jiyang.yu, jingen.liu, liefeng.bo, tmei\}@jd.com}
%\and
}

\maketitle
\ificcvfinal\thispagestyle{empty}\fi

%%%%%%%%% ABSTRACT
\begin{abstract}
   In this paper, we propose a novel video super-resolution method that aims at generating high-fidelity high-resolution (HR) videos from low-resolution (LR) ones. 
   Previous methods predominantly leverage temporal neighbor frames to assist the super-resolution of the current frame. 
   Those methods achieve limited performance as they suffer from the challenge in spatial frame alignment and the lack of useful information from similar LR neighbor frames.
   In contrast, we devise a cross-frame non-local attention mechanism that allows video super-resolution without frame alignment, leading to be more robust to large motions in the video.
%   Since the neighbor frames are highly similar to each other, the benefit from referring to the current video is limited.
%   To utilize the information beyond the current video that is being super-resolved, we design a novel memory-augmented attention module to memorize details from previously trained video samples.
   In addition, to acquire the information beyond neighbor frames, we design a novel memory-augmented attention module to memorize general video details during the super-resolution training.
   Experimental results indicate that our method can achieve superior performance on large motion videos comparing to the state-of-the-art methods without aligning frames.
   Our source code will be released.
\end{abstract}

%%%%%%%%% BODY TEXT
\input{Introduction}

\input{Relatedwork}
\input{Methodology}
\input{Experiments}
\input{Conclusion}
%-------------------------------------------------------------------------

%------------------------------------------------------------------------

{\small
\bibliographystyle{ieee_fullname}
\bibliography{egbib}
}

%\ifarXiv
%    \foreach \x in {1,...,\numbersupplementpages}
%    {
%        \clearpage
%        \includepdf[pages={\x,{}}]{\supplementfilename}
%    }
%\fi


\section*{Supplementary material}

# Supplementary Material

In the main paper, we show that our method can generate superior super-resolution results for parkour videos captured by egocentric sports cameras. In addition to the parkour videos, we collect real-world video clips (See Fig. 1) from commonly seen categories: animation, movie, music video (MV) and vlog. The goal of this experiment is to test the robustness of our method in various scenarios. Note that the appearances of these videos are significantly different from the videos in our training set (Vimeo90K [4]). In this supplementary material, the results show that our method generalizes better than comparison methods EDVR [3], TOFlow [4] and TGA [1] which are also trained on Vimeo90K [4]. We now discuss the examples shown in Fig. 1.

**1) Animation.** Animation is challenging to frame alignment for its lack of textures and low frame rate. We show a 2D animation example (a) and 3D animation examples (b) and (c) in Fig. 1. In example (a), it is difficult to recover the facial details by simply fusing neighbor frames since the information is completely corrupted in the low-resolution frames (see bicubic and comparison results). Our method can recover the facial details thanks to the memory-augmented attention. In example (b), aligning the strings on a moving guitar is difficult for EDVR [3], TOFlow [4] and TGA [1]. Our one-hot cross-frame non-local attention can effectively avoid the artifacts caused by the misaligned frames. This mechanism also works better for still repetitive pattern regions like example (c), which are often recognized as moving patterns and shifted (EDVR [3], TOFlow [4] and TGA [1]). Image super-resolution method CSNLN [2] also fails due to the erroneous non-local attention.

**2) Movie.** Super-resolving movies are of interest in online video streaming services. Movies are also challenging for their large motion and low illumination. Our method can generate very small scale sharp details like the star on the shield (examples (d)), wrinkles on the face (example (e)) and eagle eye/feathers (example (f)), which are difficult to reconstruct using either frame alignment (EDVR [3], TOFlow [4] and TGA [1]) and regular non-local attention (PFNL [5] and CSNLN [2]).

**3) MV.** Music video (MV) is one of the most-watched video categories online. Music video usually focuses on close-up shots of dancing humans, making the reconstruction of details on clothes important. In examples (g) and (h), our cross-frame non-local attention module enables recovering the fine details under the presence of motion blur. The comparison non-lcoal attention based methods PFNL [5] and CSNLN [2] cannot achieve results comparable to ours since it is much less efficient to query a pixel in the entire video segment/frame than in the pixel's neighborhood like our method.

**4) Vlog.** Vlog is another type of daily video captured by hand-held devices. This category also covers video chat, which is also common in daily life. Due to the instability of hand-held devices, these videos are extremely shaky and difficult to super-resolve. Existing video super-resolution methods TOFlow [4], TGA [1] and PFNL [5] fail in example (j) and (k). EDVR [3] can recover some details in examples (i), (j) and (k), but their results are blurry in general due to the inaccurate frame alignment.

|        | PSNR  | SSIM   | LPIPS  |       | PSNR  | SSIM   | LPIPS  |
|--------|-------|--------|--------|-------|-------|--------|--------|
| **MANA** | **37.31** | **0.9614** | **0.0681** | TGA   | 37.12 | 0.9601 | 0.0707 |
| EDVR   | 34.48 | 0.9456 | 0.1150 | PFNL  | 37.04 | 0.9673 | 0.0819 |
| TOFlow | 35.58 | 0.9531 | 0.0968 | CSNLN | 36.09 | 0.9545 | 0.0844 |

Table 1: Quantitative comparison on the videos shown in Fig. 1. Larger numbers indicate better results for PSNR and SSIM, smaller numbers indicate better results for LPIPS.

We also show the average quantitative values for the example videos in Fig. 1. In summary, our method works consistently better than the existing state-of-the-art video super-resolution methods in various categories of real-world videos. This proves the robustness of our method, which is important for real applications.

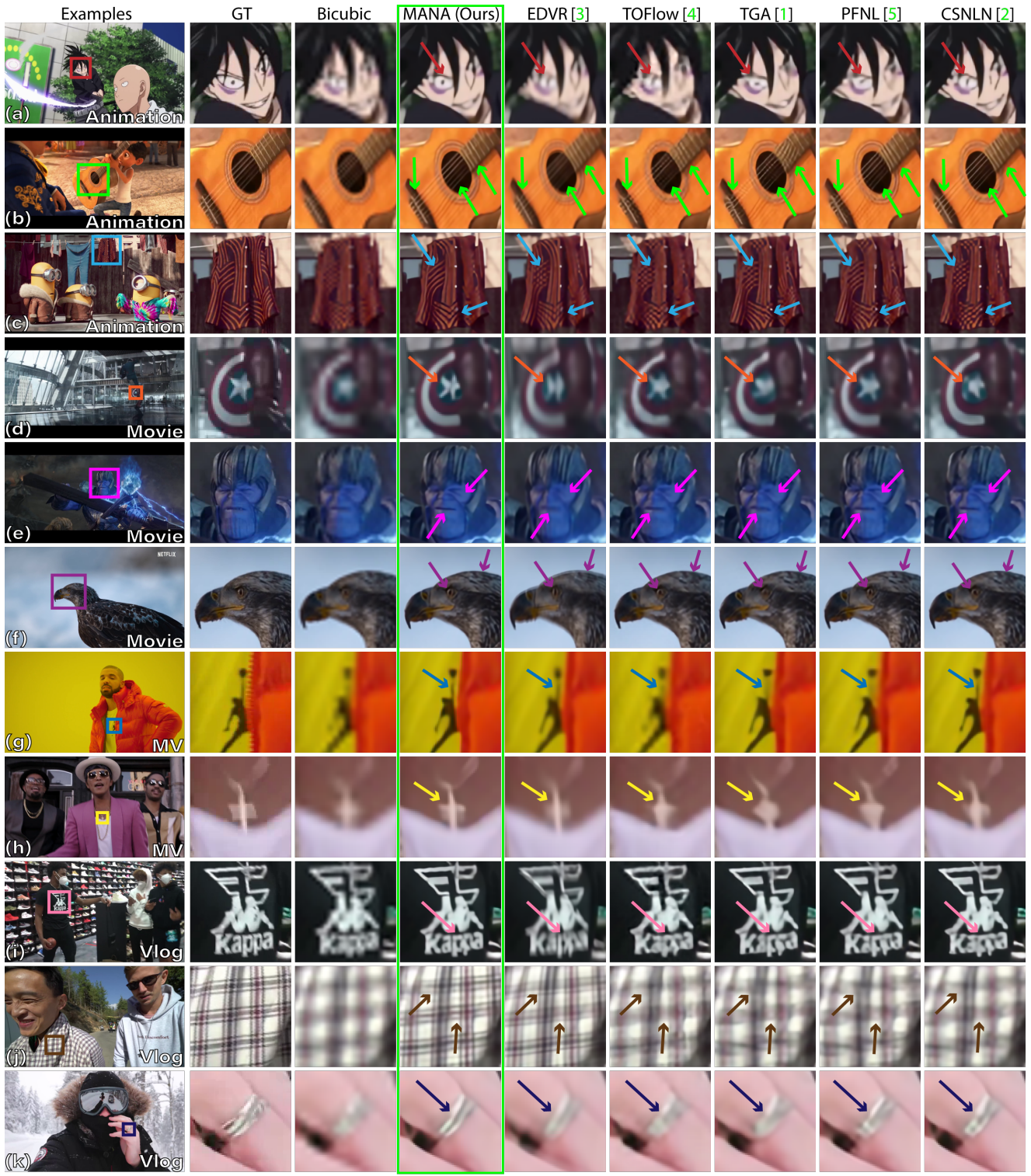

Figure 1: Additional visual comparison on examples from daily videos including animations (examples (a), (b) and (c)), movies (examples (d), (e) and (f)), MVs (examples (g) and (h)), and vlogs (examples (i), (j) and (k)). Our method works consistently better in common types of real-world video, indicating the robustness of our method.


\end{document}

%% file: Introduction.tex
\begin{figure}[htb]
    \centering
    \includegraphics[width=0.45\textwidth]{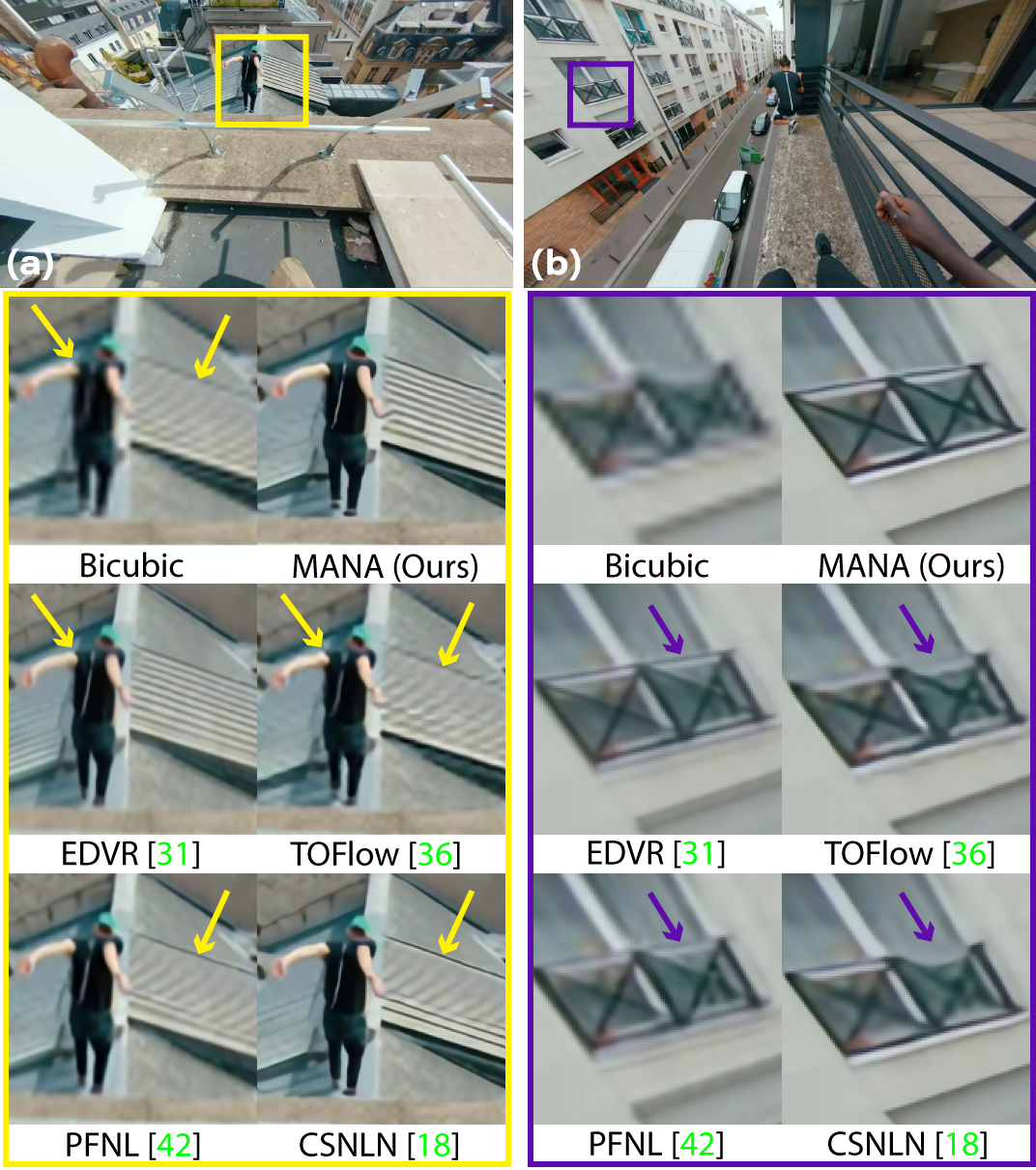}
    \caption{Our memory-augmented cross-frame non-local attention approach is robust to large motion videos (first row). Our method reconstructs visually pleasing details on repetitive patterns (left example) and thin structures (right example) while the state-of-the-art super-resolution methods requiring frame alignment (EDVR~\cite{edvr} and TOFlow~\cite{vimeo90k}) and regular non-local attention (PFNL~\cite{pfnl} and CSNLN~\cite{mei}) fails in these cases.}
    \label{fig:teaser}
    \vspace{-15pt}
\end{figure}
\vspace{-15pt}
\section{Introduction} \label{sec:intro}
Video super-resolution task aims to generate high-resolution videos from low-resolution input videos and recover high frequency details in the frames.
It is attracting more attention due to its potential application in online video streaming services and the movie industry.

There are two major challenges in the video super-resolution tasks.
The first challenge comes from the dynamic nature of videos.
To ensure temporal consistency and improve visual fidelity, video super-resolution methods seek to fuse information from multiple neighbor frames.
Due to the motion across the frames in the video, neighbor frames need to be aligned before fusion.
Recent video super-resolution works have proposed various ways for aligning neighbor frames to the current frame, either by explicit warping using optical flow~\cite{caballero, liu,frvsr,spmc} or learning implicit alignment using deformable convolution~\cite{tdan, edvr}.
However, the quality of these works highly depends on the accuracy of spatial alignment of neighbor frames, which is difficult to achieve in videos with large motions.
This hinders the application of existing video super-resolution methods in real-world videos such as egocentric sports videos (see our Parkour dataset in Sec.~\ref{subsec:dataset}), and some videos from animation, movies and vlogs (see additional examples in supplementary material). 
%e.g. egocentric sports videos.
%To evaluate the robustness of the video super-resolution methods, we collected a dataset called \textit{Parkour dataset} containing videos with large motions. (See Sec.~\ref{subsec:dataset})
%In the supplementary material, we provide additional examples from other common categories of real-world videos, e.g. animation, movies, and vlogs.

The second challenge comes from the irreversible loss of high-frequency detail and the lack of useful information in the low-resolution video.
Recent learning based single image super-resolution works~\cite{dongetal, kim1, kim2,ledig,mei, sun,timofte,wangetal,yang,adobe} have intensively studied the visual reconstruction from low-resolution images by learning general image prior to help recover high-frequency details or transferring texture from a high-resolution reference image.
%studied the visual detail reconstruction in low-resolution images intensively~\cite{dongetal, kim1, kim2,ledig,mei, sun,timofte,wangetal,yang,adobe} by learning general image prior to help recover high-frequency details or transferring texture from a high-resolution reference image.
Since these methods do not guarantee temporal consistency in the visual appearance, they usually generate results inferior to that of video super-resolution methods using neighbor frame information.
However, in the video super-resolution task, the neighbor frames are largely similar to each other and the benefits from fusing them are limited.
For large motion videos, the neighbor frames become less similar.
In this scenario, the correlation among neighbor frames also becomes smaller, and the video super-resolution essentially degrades to the single image super-resolution since it cannot find any useful information by mining neighbor frames.

To address these issues, we propose a memory-augmented non-local attention framework for video super-resolution.
Our method is a deep learning based method.
%The input of our network is a set of consecutive low-resolution video frames.
Taking a set of consecutive low-resolution video frames as inputs, our network produces the high-resolution version of the temporal center frame by referring to the information from its neighbor frames.
Since consecutive frames share a large portion of visual contents, this scheme implicitly ensures the temporal consistency in the result.

To solve the frame-alignment challenge, we design the \textit{Cross-Frame Non-local Attention} module which allows us to fuse neighbor frames without aligning them towards the current frame.
Although conventional non-local attention can capture temporally and spatially long-distance correspondences, it requires computing pair-wise correlation between each pixel in the query and key.
This imposes a large burden on GPU memory since in the video super-resolution case down-sampling the video like Wang et al.~\cite{nonlocal} and losing more high-frequency detail is not desired.
To make non-local attention practical in video super-resolution, in the cross-frame non-local attention module, we only query the current frame pixel within its $9 \times 9$ spatial neighborhood in the neighbor frames.
Furthermore, instead of using softmax normalized correlation matrix to combine the value tensor like it is done in traditional non-local attention, we only sample the most correlated pixel in the value tensor, namely \textit{one-hot attention}.
%In Sec.~\ref{sec:result}, we will show that our cross-frame non-local attention scheme greatly improves the results in large motion videos.
Our one-hot non-local attention is effective especially for videos with large motions. 
In Fig.~\ref{fig:teaser}(a), while the state-of-the-art video super-resolution method EDVR~\cite{edvr} and TOFlow~\cite{vimeo90k} fails due to fusing misaligned frames, our method reconstruct sharp details like the stripes on the roof and the waving arm.
We provide complete verification of the effectiveness of our one-hot non-local attention framework in Sec.~\ref{sec:result}.

To solve the challenge of the lack of information, we seek to fuse useful information beyond the current video.
This means that the network should \textit{memorize} previous experiences in super-resolving other videos in the training set.
Based on this principle, we introduce a \textit{Memory-Augmented Attention} module to our network.
In this module, we maintain a 2D memory bank which is completely learned during the video super-resolution training.
The purpose of this module is to summarize the representative local details in the entire training set and use them as an external reference for super-resolving the current video frame.
To our experience, by introducing the memory bank mechanism, our work is the first video super-resolution method that incorporates information beyond the current video.
With the help of the memory-augmented attention module, our method can recover details that are missing in the low-resolution video like the balcony railings in Fig.~\ref{fig:teaser}(b).

In this paper, our contributions include the follows:

\noindent \textbf{Cross-frame non-local attention.} We introduce a novel cross-frame non-local attention that liberates the video super-resolution from the error-prone frame alignment process.
This design makes our method robust to videos with large motions.  (See Sec.~\ref{subsec:nonlocal})

\noindent \textbf{Video super-resolution beyond current video.} We proposed a novel memory-augmentation mechanism in video super-resolution, which memorizes previous experiences during the training process and uses the memory to assist current video super-resolution. (See Sec.~\ref{subsec:memory} and Sec.~\ref{subsec:detail}).

%% file: Relatedwork.tex
\section{Related Work}\label{sec:relatedwork}
%In this section, we summarize the related work in single image super-resolution and video super-resolution in general, then focus on the recent works applying attention mechanism. Finally we discuss the memory bank models that inspired our work.
\noindent \textbf{Single Image Super-Resolution}
Early image super-resolution works resort to image processing algorithms~\cite{sunjian,jianchao3,jianchao2,jianchao}.
Recent works in deep learning have been proved to obtain superior results in image super-resolution due to the ability to learn prior of high-resolution images.
SRCNN proposed by Dong et al.~\cite{dongetal} first introduces a convolutional neural network in image super-resolution.
Kim et al. further develop VDSR~\cite{kim1} and DRCN~\cite{kim2} and explore deeper residual networks and recursive structures.
ESPCN~\cite{espcn} encode the low-resolution image into multiple sub-pixel channels and upscale to a high-resolution image by shuffling the channels back in the spatial domain.
This idea was widely used in recent super-resolution works.
Other approaches using CNN includes pyramid strucutre (LapSRN~\cite{lapsrn}), recursive residual network (DRRN~\cite{tai}), dense skip connections (SRDenseNet~\cite{densenet} and RDN~\cite{rdn}), and adversarial networks (SRGAN~\cite{ledig}, EnhanceNet~\cite{enhancenet}, ESRGAN~\cite{esrgan} and GLEAN~\cite{glean}).

\noindent \textbf{Video Super-Resolution}
Video super-resolution typically generate better result than single image super-resolution thanks to the extra information from neighbor frames.
The main focus of video super-resolution works is how to correctly fuse auxiliary frames in the presence of dynamic contents and camera motion.
Some methods explicitly use optical flow (VESPCN~\cite{caballero}, FRVSR~\cite{frvsr}, SPMC~\cite{spmc}, TOFlow~\cite{vimeo90k} and BasicVSR/IconVSR~\cite{basicvsr}) or homography (TGA~\cite{tga}) to align neighbor frames.
However, estimating accurate optical flow/transformation is challenging when the motion between the neighbor frame and current frame is large.
Having observed this limitation, recent methods start to explore techniques to bypass alignment or implicitly align frames.
Jo et al. proposed DUF~\cite{duf} that learns dynamic upsampling filters that combine the entire spatial neighborhood of a pixel in the auxiliary frames.
TDAN~\cite{tdan} and EDVR~\cite{edvr} use deformable convolution layer to sample neighbor frames according to the estimated kernel offsets.
However, these methods essentially still learn the spatial correspondence across frames.
As we will show in Sec.~\ref{sec:result}, in large motion cases, the results from these methods are unsatisfactory.
Unlike any previous video super-resolution methods, our method finds the pixel correspondence in an unstructured fashion by applying non-local attention.

\noindent \textbf{Non-local Attention in Super-Resolution}
Attention mechanism has proven to be effective in various computer vision tasks\cite{iclr_transformer, sqe, standalone, sttn,rcan,yulun}.
Non-local neural networks proposed by Wang et al.\cite{nonlocal} capture pixel-wise correlations within a video segment, making temporally and spatially long distance attention possible.
Recent image super-resolution methods using non-local attention includes CSNLN~\cite{mei}, RNAN~\cite{yulun} and TTSR~\cite{ttsr}.
Video super-resolution method PFNL~\cite{pfnl} also utilize self-attention over a set of consecutive video frames.
However, directly applying non-local attention requires storing pair-wise correlation between query and key.
In the video super-resolution task, the size of the correlation matrix grows quadratically with the total number of pixels in the video segment and becomes intractable when the input frame size is large.
Moreover, a larger number of pixels will potentially degrade the performance of non-local attention as we will discuss in Sec.~\ref{subsec:nonlocal}.
Our work performs one-hot non-local attention within the patch enclosing a query pixel and only selects the most correlated pixel in the neighbor frames.
This approach greatly reduced the GPU memory usage and generate better results comparing to that of PFNL~\cite{pfnl}.

\begin{figure}[t]
    \centering
    \includegraphics[width=0.45\textwidth]{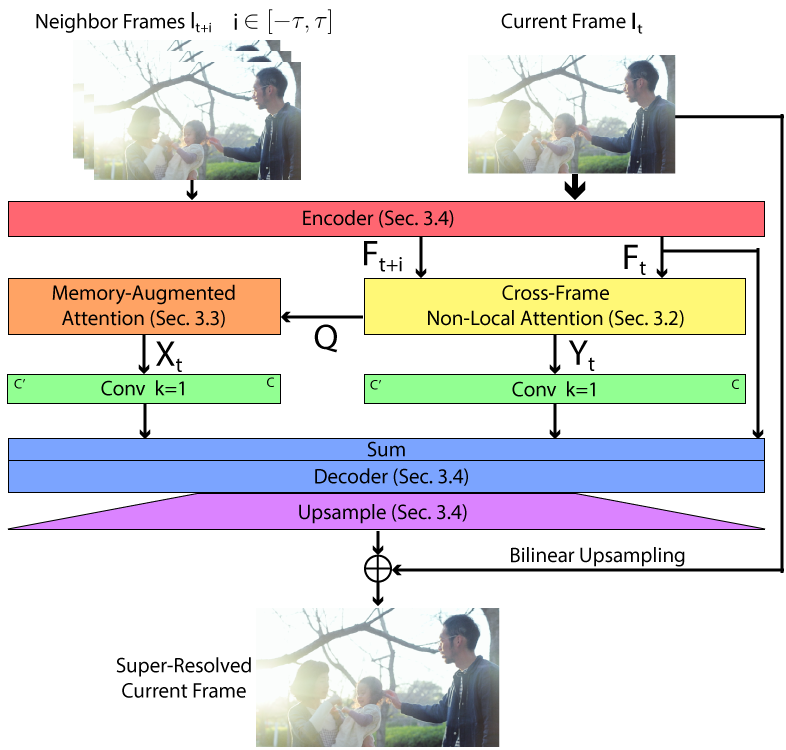}
    \caption{An overview of the structure of our video super-resolution network. The network super-resolves the current frame $\mathbf{I}_t$ using the neighbor frames ${\mathbf{I}_{t-\tau},...,\mathbf{I}_{t+\tau}}$ as the input. The cross-frame non-local attention aims at mining information from neighbor frames and the memory-augmented attention targets at memorizing experience in super-resolving other videos. The output of these modules are used as residual to enhance the details of a bilinearly upsampled low-resolution frame. }
    \label{fig:pipeline}
    \vspace{-15pt}
\end{figure}

\noindent \textbf{Memory models}
Neural networks with memory show their potential in natural language processing~\cite{nlpmem,membank}, image classification~\cite{memclass} and video action recognition~\cite{memdpc}.
These works augment their model with an explicit memory bank that can be updated or read during the training.
Inspired by these works, we design a memory-augmented attention module to incorporate previous knowledge gained from super-resolving other videos.
In Sec.~\ref{sec:result}, we will show that the memory module provides a significant boost in the performance of video super-resolution.

%% file: Methodology.tex
\section{Methodology}
%In this section, we present our approach to the videos super-resolution problem.
%We first discuss the principle of our approach and provide an overview of our network in Sec.~\ref{subsec:overview}.
%In Sec.~\ref{subsec:nonlocal} and Sec.~\ref{subsec:memory}, we discuss the details of the building blocks in our network, namely \textit{Cross-Frame Non-local Attention} and \textit{Memory-Augmented Attention}, which are responsible for fusing the information from the current video and general videos respectively.
%Finally, we introduce the parameters of our network and the training procedure in Sec.~\ref{subsec:detail}.

\subsection{Overview}\label{subsec:overview}
Fig.~\ref{fig:pipeline} demonstrates the structure of our video super-resolution network.
The goal of our network is to super-resolve a single low-resolution frame $\mathbf{I}_{t} \in \mathbb{R}^{3\times H \times W}$, given the low-resolution temporal neighbor frames $\left \{\mathbf{I}_{t-\tau},...,\mathbf{I}_{t+\tau}\right \}$, where $H$ and $W$ are the video height and width respectively.
To make the discussion more concise, we will use ``current frame" to refer to $\mathbf{I}_{t}$ and ``neighbor frames" to refer to $\left \{\mathbf{I}_{t-\tau},...,\mathbf{I}_{t+\tau}\right \}$. 
We will use $T=2\tau+1$ to represent the time span of neighbor frames.
Note that neighbor frames include the current frame.

The first stage of our network embeds all the video frames into the same feature space by applying the same encoding network to each input frame.
We denote the embedded features as $\left \{\mathbf{F}_{t-\tau},...,\mathbf{F}_{t+\tau}\right \}\in \mathbb{R}^{C \times H \times W}$, where $C$ is the dimension of the feature space.

As we discussed in Sec.~\ref{sec:intro}, our super-resolution process refers to both the current video and general videos.
Based on this principle, we adapt the attention mechanism which allows us to query the pixels that need to be super-resolved in the keys consist of auxiliary pixels. 
Specifically, the second stage of our network includes two parts: \textit{Cross-Frame Non-local Attention} and \textit{Memory-Augmented Attention}.

\textit{Cross-Frame Non-local Attention} aims to mine useful information from neighbor frame features.
In this module, neighbor frame features are queried by the current frame feature and the most possible match will be selected as the output.
We denote the output of the cross-frame non-local attention module as $\mathbf{X}_{t}\in \mathbb{R}^{C' \times H \times W}$, where $C'=C/2$ is the dimension of the embedding space of the cross-frame non-local attention module.
The design of this module will be discussed in Sec.~\ref{subsec:nonlocal}.

\textit{Memory-Augmented Attention} maintains a global memory bank $\mathbf{M}\in \mathbb{R}^{C' \times N}$ to memorize useful information from general videos in the training set, where $N$ represents an arbitrary number of entries in the memory bank. 
We use the current frame feature to query the memory bank directly.
However, unlike the cross-frame non-local attention module in which the keys are embedded versions of neighbor frame features, the memory bank is completely learned.
The output of this module is denoted as $\mathbf{Y}_{t}\in \mathbb{R}^{C' \times H \times W}$.
This module will be discussed in Sec.~\ref{subsec:memory}.

Finally, the output of the cross-frame non-local attention module $\mathbf{X}_{t}$ and memory-augmented attention module $\mathbf{Y}_{t}$ are convolved by two different convolutional layers with kernel size 1 and added to the input current frame feature $\mathbf{F}_t$ as residuals.
A decoder decodes the output of attention modules and an up-sampling module shuffles the pixels to generate a high-resolution residual.
The residual adds details to the bilinearly up-sampled blurry low-resolution frame, resulting in a clear high-resolution frame.

\begin{figure}[t]
    \centering
    \includegraphics[width=0.45\textwidth]{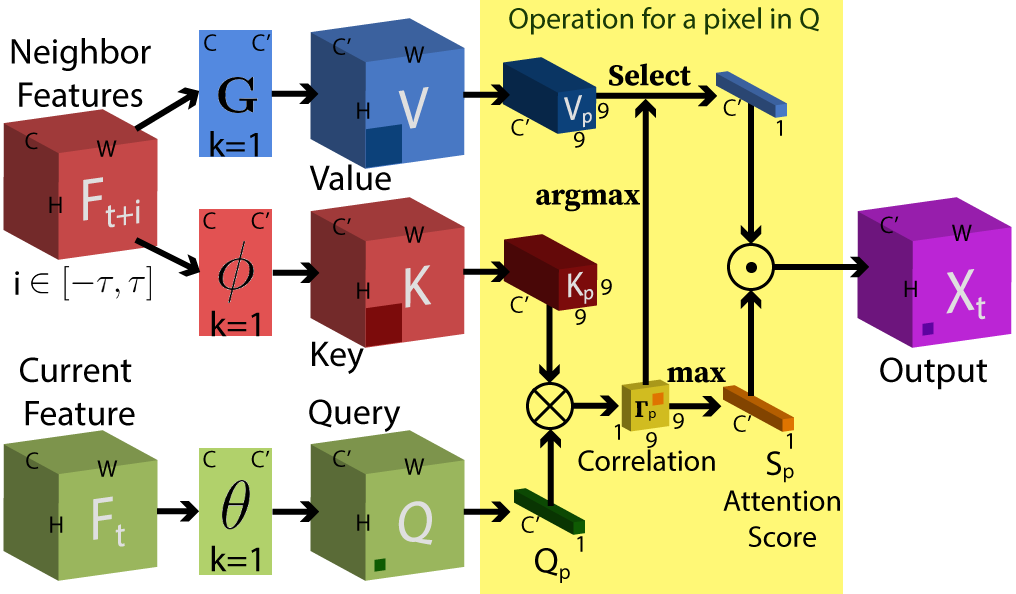}
    \caption{The cross-frame non-local attention module in our network. The size are marked on the edges of the tensors. The operation marked by the yellow box is done in parallel for each pixel $\mathbf{Q}_p$ in the query tensor $\mathbf{Q}$. Best viewed in PDF.}
    \label{fig:nonlocal}
    \vspace{-10pt}
\end{figure}

\subsection{Cross-Frame Non-local Attention}\label{subsec:nonlocal}
One of the major procedures in the conventional video super-resolution methods is to align the neighbor frames so that the corresponding pixels can be fused and improve the quality of the super-resolution of the current frame.
To achieve the alignment, the typical approaches in video super-resolution works include optical flow~\cite{frvsr,vimeo90k} and deformable convolution~\cite{tdan,edvr}.
However, aligning pixels according to color consistency is known to be a challenging task under large motion or illumination change.
As a consequence, the inaccuracy in alignment will negatively impact the performance of video super-resolution.
In our work, we seek to avoid this performance overhead.
As we discussed in Sec.~\ref{sec:relatedwork}, the non-local attention~\cite{nonlocal} enables capturing temporally and spatially long distance correspondence.
Therefore, the frame alignment can be omitted if non-local attention is used to query pixels of the current frame in neighbor frames.

The cross-frame non-local attention module is demonstrated in Fig.~\ref{fig:nonlocal}.
We first normalize the input frame features using group normalization~\cite{groupnorm}, resulting in the normalized neighbor frame features $\left \{  \mathbf{\overline{F}}_{t-\tau},...,\mathbf{\overline{F}}_{t+\tau}\right \}$.
In our non-local attention setup, the center feature $\mathbf{\overline{F}}_{t}$ is used as the query tensor, and neighbor frame features $\left \{  \mathbf{\overline{F}}_{t-\tau},...,\mathbf{\overline{F}}_{t+\tau}\right \}$ serve as both the key and value tensors.
The embedded version of query, key and value tensor are noted as $\mathbf{Q} \in \mathbb{R}^{C'\times H \times W}$, $\mathbf{K} \in \mathbb{R}^{C'\times T\times H \times W }$ and $\mathbf{V} \in \mathbb{R}^{C'\times T\times H \times W}$ in Fig.~\ref{fig:nonlocal}.
In the traditional setup of non-local attention, the next step is to flatten the temporal and spatial dimension of $\mathbf{Q}$ and $\mathbf{K}$ to $\mathbf{\widehat{Q}} \in \mathbb{R}^{HW\times C'}$ and $\mathbf{\widehat{K}} \in \mathbb{R}^{C'\times HWT}$ and calculating the correlation matrix $\mathbf{\Gamma}=\mathbf{\widehat{Q}}\mathbf{\widehat{K}}$.
However, the size of $\mathbf{\Gamma}$ is $HW \times HWT$.
Even though the input to our network is low-resolution frames, the size of this matrix is large and grows quadratically when involving more neighbor frames.
This makes the training difficult due to the limited GPU memory.
Moreover, note that the number of columns in $\mathbf{\Gamma}$ depends on the frame size.
With the same pre-trained network, normalizing $HWT$ dimensions using softmax makes the columns in $\mathbf{\widehat{K}}$ that are most correlated with a row in $\mathbf{\widehat{Q}}$ less significant when the input frame size is larger.
Intuitively, traditional non-local attention weakens the contribution of the most useful auxiliary pixels in the super-resolution case.
In Sec.~\ref{subsec:quant}, we will show that traditional non-local attention did not benefit the video super-resolution method PFNL~\cite{pfnl} which directly applies it to the entire group of neighbor frames.

To mitigate the GPU memory issue, we conduct non-local attention on each neighbor frames separately.
Moreover, for each pixel in the query tensor, we conduct non-local attention only on its spatial neighbor region in the temporal neighbor frames.
The insight is that video motion is limited during a short time period, and a pixel's correspondences in the neighbor frames lie in the neighborhood of its position.
Specifically, we unfold each temporal slice of $\mathbf{K}$ with dimension $C'\times H \times W$ into $9\times9$ patches in the spatial domain.
Therefore, for each pixel $\mathbf{\widehat{Q}}_p \in \mathbb{R}^{1 \times C'}$ in the query tensor $\mathbf{\widehat{Q}}$, the corresponding key tensor becomes $\mathbf{K}_p \in \mathbb{R}^{C' \times 81}$.
As a result, the correlation matrix $\mathbf{\Gamma}_{p}=\mathbf{\widehat{Q}}_p\mathbf{K}_p$ has a dimension of $1 \times 81$.
Note that $\mathbf{K}_p$ is different for each row of $\mathbf{\widehat{Q}}$; each $\mathbf{K}_p$ only represents the spatial neighborhood of the specific pixel in the current frame.
By stacking the $\mathbf{\Gamma_{p}}$ for each pixel, we obtain the final correlation matrix $\mathbf{\Gamma}$.

To resolve the performance degradation issue in the traditional non-local attention, we reduce the non-local attention into an \textit{one-hot} attention.
Specifically, for each row in $\mathbf{\Gamma}$, we only select the largest entry as follows:
\begin{equation}
%\begin{aligned}
    s_i=\max_{j}{\mathbf{\Gamma}(i,j)},\quad d_i=\argmax_{j}{\mathbf{\Gamma}(i,j)}
%\end{aligned}
\end{equation}
where $s_i$ is the attention score and $d_i$ is the column coordinate in $\mathbf{K}_p$.
Denote the \textit{one-hot} attention results as $\mathbf{S} \in \mathbb{R}^{HW\times 1}$ and $\mathbf{D} \in \mathbb{R}^{HW\times 1}$ which consists of $s_i$ and $d_i$ respectively.
We then select the value tensor $\mathbf{V}_p$ as follows:
\begin{equation}\label{eqn:select}
\mathbf{\widehat{X}}_t=\mathbf{S} \cdot  \mathbf{V}_p(\mathbf{D})
\end{equation}
In Eqn.~\ref{eqn:select}, $\mathbf{V}_p \in \mathbb{R}^{C' \times 81}$ is the value tensor unfolded in the same way as $\mathbf{K}_p$.
The operator $(\ \ )$ selects $HW$ columns from $\mathbf{V}_p$ using the $HW$ indices in $\mathbf{D}$.
Then the selected $HW$ columns are multiplied by $HW$ attention scores in $\mathbf{S}$.
The $\mathbf{\widehat{X}}_t \in \mathbb{R}^{C' \times HW}$ is reshaped to $\mathbf{X}_t \in \mathbb{R}^{C' \times H \times W}$ as the output of the cross-frame non-local attention module.

\begin{figure}[t]
    \centering
    \includegraphics[width=0.45\textwidth]{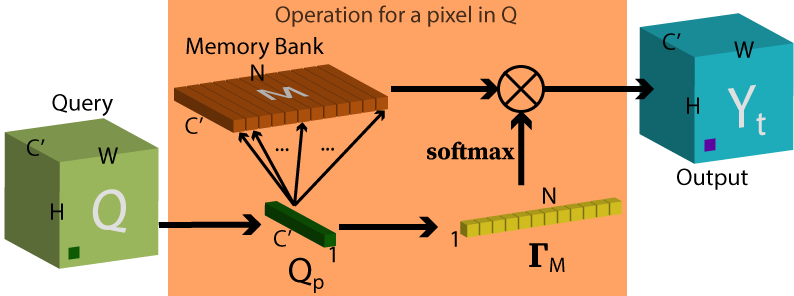}
    \caption{The memory-augmented attention module in our network. The operation marked by the orange box is done in parallel for each pixel $\mathbf{Q}_p$ in the query tensor $\mathbf{Q}$. Best viewed in PDF.}
    \label{fig:memory}
    \vspace{-10pt}
\end{figure}

\subsection{Memory-Augmented Attention}\label{subsec:memory}
Cross-frame non-local attention enables the fusion of the information from neighbor frames in the current video.
However, the neighbor frames used in the attention are also low-resolution with similar content to the current frame.
Therefore, the benefit from cross-frame non-local attention is limited.
We seek to refer to more local detail information beyond the current video, which requires memorizing useful information from the entire training set.
For this purpose, our network includes a memory-augmented attention module.
The module maintains a global memory bank $\mathbf{M} \in \mathbb{R}^{C' \times N}$ which is learned as parameters of the network.
We use regular non-local attention to query current frame features $\mathbf{\widehat{Q}}$ in the global memory bank $\mathbf{M}$, i.e. the correlation matrix is $\mathbf{\Gamma}_M= \mathbf{\widehat{Q}}\mathbf{M} \in \mathbb{R}^{HW\times N}$.
Finally, we obtain the output
\begin{equation}
    \mathbf{\widehat{Y}}_t=softmax(\mathbf{\Gamma}_M) \mathbf{\widehat{M}}
\end{equation}
where $\mathbf{\widehat{M}} \in \mathbb{R}^{N \times C'}$ is the transposed version of the memory bank $\mathbf{M}$.
Similar to the cross-frame non-local attention module, we reshape $\mathbf{\widehat{Y}}_t \in \mathbb{R}^{HW \times C'}$ to $\mathbf{Y}_t \in \mathbb{R}^{C' \times H \times W}$ as the output of the memory-augmented attention module.

\subsection{Implementation Details}\label{subsec:detail}
\noindent \textbf{Training Set} 
Vimeo90K is a large-scale video dataset proposed by Xue et al.~\cite{vimeo90k}. Following recent super-resolution methods TOFlow~\cite{vimeo90k}, TDAN~\cite{tdan} and EDVR~\cite{edvr}, we use the training set of Vimeo90K to train our network.
Each video clip in Vimeo90K consists of 7 consecutive frames.
We use the center frame as the current frame to be super-resolved.
All 7 frames are used as the neighbor frames.

\begin{figure}[t]
    \centering
    \includegraphics[width=0.45\textwidth]{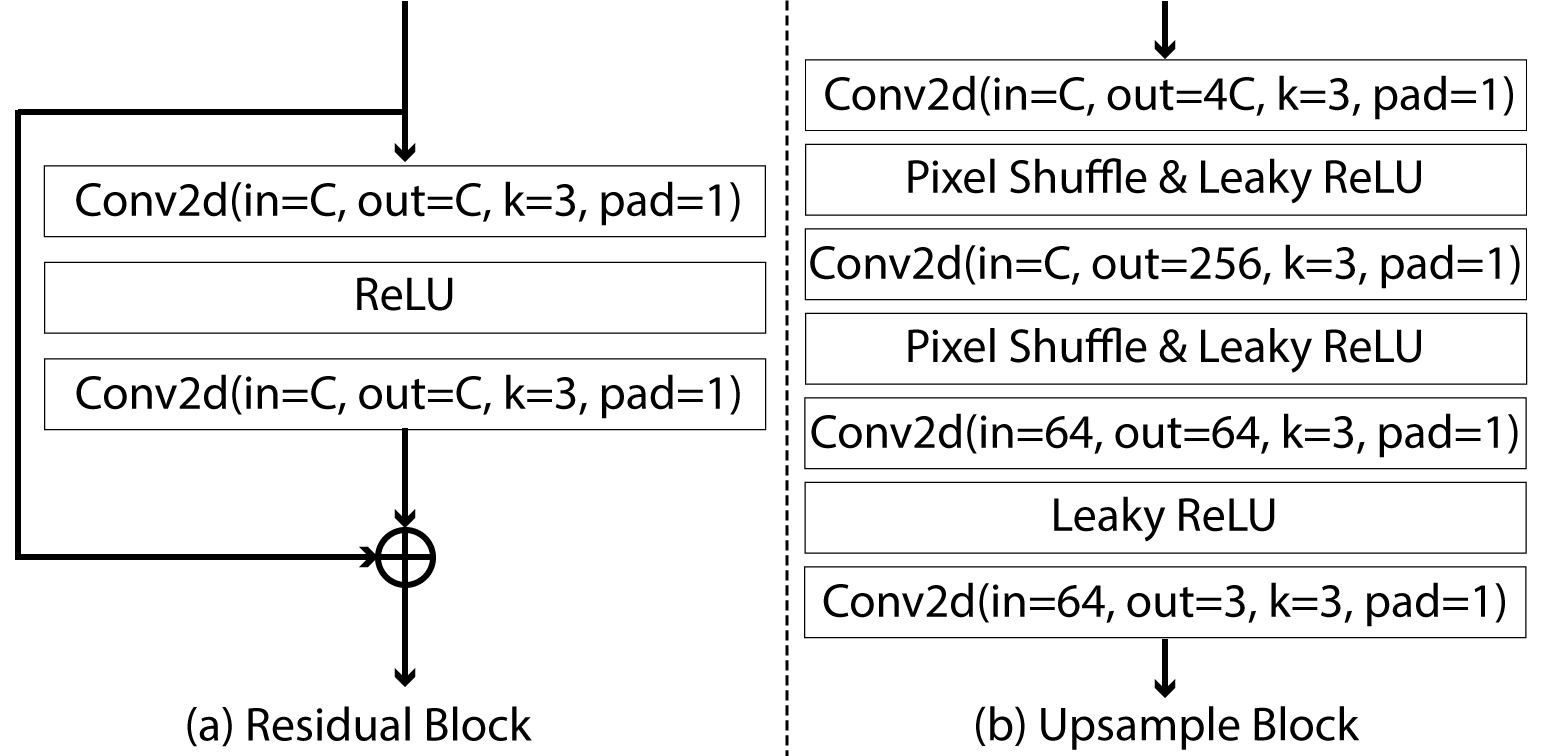}
    \caption{Basic building blocks in our network. (a) Residual blocks are used to build the encoder and decoder. (b) Upsample block shuffles pixels in different channels into a high-resolution frame.}
    \label{fig:blocks}
    \vspace{-10pt}
\end{figure}

\noindent \textbf{Network Structure}
Besides the structures of cross-frame non-local attention and memory-augmented attention module shown in Fig.~\ref{fig:pipeline}, we demonstrate the structure of other basic building blocks in Fig.~\ref{fig:blocks}. 
The residual blocks (Fig.~\ref{fig:blocks}(a)) are used to build the frame encoder and decoder.
The frame encoder and decoder are the concatenation of 5 residual blocks and 40 residual blocks respectively.
The structure of the up-sampling block is shown in Fig.~\ref{fig:blocks}(b).
In this paper, we focus on 4x video super-resolution task.
The up-sampling block is built by 2 pixel shuffle blocks, each up-sample the feature map by 2 using the pixel shuffle operation defined in ESPCN~\cite{espcn}.
We use $C=128$ for all experiments in this paper.

\begin{figure}[t]
    \centering
    \includegraphics[width=0.45\textwidth]{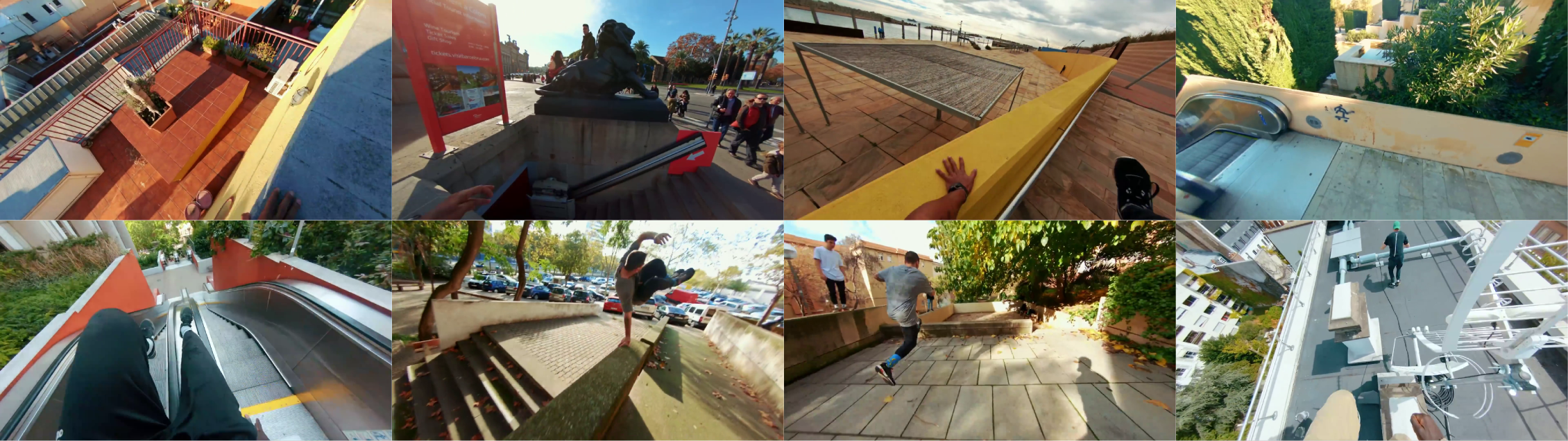}
    \caption{Video stills from the \textit{Parkour dataset}. Due to the large camera motion in this dataset, it is challenging for exisitng video super-resolution methods.}
    \label{fig:dataset}
    \vspace{-10pt}
\end{figure}

\noindent \textbf{Training Procedure} 
We implement our network in PyTorch~\cite{pytorch} and use Adam optimizer~\cite{adam} with $\beta_1=0.5$ and $\beta_2=0.99$ for training.
The weight of the last convolutional layers of the cross-frame non-local attention module and the memory-augmented attention module is initialized to zero.
The training of our network consists of three stages.

In the first stage, we fix the memory-augmented attention module and train the rest part of the network for 90,000 iterations at the learning rate of $10^{-4}$.
The loss function used is $L_1=\left \|  \mathbf{O}_t-\mathbf{G}_t\right \|_1$, where $\mathbf{O}_t$ stands for the output super-resolved current frame and $\mathbf{G}_t$ is the ground truth high-resolution frame.

In the second stage, we fix the network weights except for the memory-augmented attention module.
The loss function $L_2=\left \|  \mathbf{Y}_t-\mathbf{Q}\right \|_1$ focus on training the memory bank.
Note that the training process optimizes the memory bank $\mathbf{M}$ so that a query $\mathbf{Q}$ can be represented by the combination of the columns in $\mathbf{M}$ as accurate as possible.
This is essentially clustering and summarizing the most representative general pixel features in the encoded space.
We train this stage for 30,000 iterations at the learning rate of $10^{-4}$.

In the final stage, we fine-tune the entire network using $L_1$ for 30,000 iterations at the learning rate of $10^{-5}$.

%% file: Experiments.tex
\begin{figure*}[t]
    \centering
    \includegraphics[width=\textwidth]{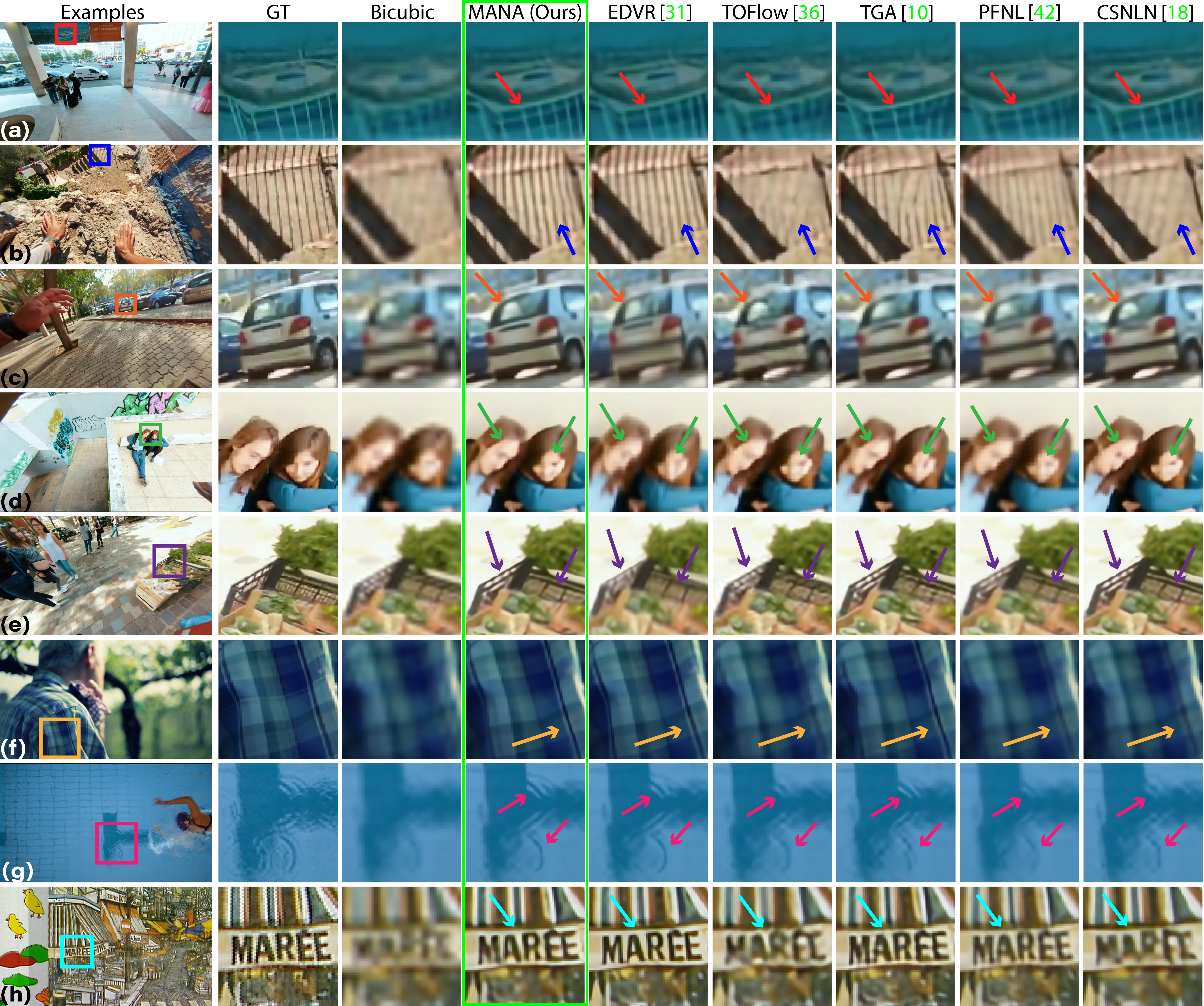}
    \caption{Visual comparison on the Parkour dataset, Vimeo90K~\cite{vimeo90k} dataset and Vid4~\cite{frvsr} dataset. Example (a), (b), (c), (d) and (e) are selected from the large motion Parkour dataset. Example (f) and (g) are selected in the Vimeo90K~\cite{vimeo90k} dataset. Example (h) is from Vid4~\cite{frvsr} dataset. We mark the inset locations on the video stills on the left. To make our discussion clearer, we add arrows pointing to the region that we will be discussing in Sec.~\ref{subsec:visual}. \textcolor{red}{Best viewed in PDF.}}
    \label{fig:visual}
    \vspace{-10pt}
\end{figure*}

\section{Experiments}\label{sec:result}
In this section, we compare our work with recent state-of-the-art video super-resolution (VSR) and single image super-resolution (SISR) methods.
We select comparison methods based on their approaches to the super-resolution problem: VSR via explicit frame alignment (TOFlow~\cite{vimeo90k} and TGA~\cite{tga}), VSR via implicit frame alignment (EDVR~\cite{edvr}), VSR via regular non-local attention (PFNL~\cite{pfnl}) and SISR via regular non-local attention (CSNLN~\cite{mei}) applied to each video frame individually.
Similar to other VSR works, in this paper, we focus on the 4x scaling case for all the comparisons shown in this section.
To obtain the low-resolution input, we use bicubic down-sampling on the ground truth high-resolution frames.
According to our experiment, PFNL~\cite{pfnl} and TGA~\cite{tga} introduce serious aliasing artifacts to the results using bicubic down-sampled video.
To make the comparison fair, for PFNL~\cite{pfnl} and TGA~\cite{tga}, we apply Gaussian blur to the ground truth frames before down-sampling following the procedure in their papers.
Unless otherwise stated, our results shown in this section are generated with the memory size of $N=256$ in the memory-augmented attention module.
We conduct the experiment on a desktop computer with an NVIDIA 2080Ti GPU.
The average processing speed of our network is 251ms per 960x540 HR frame. 

%In Sec.~\ref{subsec:dataset}, we demonstrate the testing sets used for evaluating our method and the comparison methods.
%We present the visualization of our results and discuss our advantage against the comparison methods in Sec.~\ref{subsec:visual}.
%We provide quantitative comparisons in Sec.~\ref{subsec:quant}.
%In Sec.~\ref{subsec:ablation}, we conduct ablation study on the memory-augmented attention module and discuss the benefits of using memory bank in video super-resolution.
%Finally, we encourage the readers to watch our supplementary video for better visualization of the dynamics.
\begin{table*}[t]
\footnotesize
\begin{tabular}{p{0.08\textwidth}||p{0.048\textwidth}<{\centering}|p{0.048\textwidth}<{\centering}|p{0.048\textwidth}<{\centering}||p{0.048\textwidth}<{\centering}|p{0.048\textwidth}<{\centering}|p{0.048\textwidth}<{\centering}||p{0.048\textwidth}<{\centering}|p{0.048\textwidth}<{\centering}|p{0.048\textwidth}<{\centering}||p{0.048\textwidth}<{\centering}|p{0.048\textwidth}<{\centering}|p{0.048\textwidth}<{\centering}}
            & \multicolumn{3}{c||}{(a) Parkour Dataset} & \multicolumn{3}{c||}{(b) Vimeo90K Dataset~\cite{vimeo90k}}    & \multicolumn{3}{c||}{(c)Vid4 Dataset~\cite{frvsr}} 
            & \multicolumn{3}{c}{(d)SPMC Dataset~\cite{spmc}} \\ \hline
            & PSNR in dB          & SSIM   &LPIPS         & PSNR in dB & SSIM &LPIPS  & PSNR in dB & SSIM &LPIPS  & PSNR in dB & SSIM   &LPIPS                  \\ \hline
Bicubic     & 29.51 (+3.97)              & 0.8712    &0.3101      & 29.75 (+4.96)    & 0.8476 & 0.2948 & 22.34 (+2.81)     & 0.6131   &0.5186  &25.67 (+3.55) &0.7241 &0.4270                 \\ \hline
\textbf{MANA (Ours)} & \textbf{\textcolor{red}{33.48}}              & \textbf{\textcolor{red}{0.9356}}   &\textbf{\textcolor{red}{0.1241}}       & \textbf{34.71}     & \textbf{0.9261} &\textbf{0.1101} & \textbf{25.15}     & \textbf{0.7796}    &\textbf{\textcolor{blue}{0.2744}}       &\textbf{\textcolor{red}{29.22}} &\textbf{\textcolor{red}{0.8458}} &\textbf{\textcolor{red}{0.2119}}           \\ \hline
EDVR~\cite{edvr}        & 31.61 (+1.87)              & 0.9113    &0.1900      & \textcolor{red}{35.68 (-0.97)}      & \textcolor{red}{0.9372} &\textcolor{blue}{0.1019} & \textcolor{red}{25.79 (-0.64)}     & \textcolor{red}{0.8063}    &\textcolor{red}{0.2489}      &27.98 (+1.24) &0.8109 &0.2715            \\ \hline
TOFlow~\cite{vimeo90k}      & 32.35 (+1.13)             & 0.9197     &0.1804     & 32.96 (+1.75)     & 0.9041 &0.1451 & 24.41 (+0.74)    & 0.7435   &0.3340      &28.55 (+0.67) &0.8327 &0.2661             \\ \hline
TGA~\cite{tga}         & 31.14 (+2.34)             & 0.9033    &0.2224      & \textcolor{blue}{35.03 (-0.32)}      & \textcolor{blue}{0.9310} &\textcolor{red}{0.1013}  & \textcolor{blue}{25.36 (-0.21)}     & \textcolor{blue}{0.7949}   &0.2834      &\textcolor{blue}{29.06 (+0.16)} &\textcolor{blue}{0.8449} &0.2390             \\ \hline
PFNL~\cite{pfnl}        & 32.04 (+2.44)            & 0.9189  &0.2244        & 31.86 (+2.85)     & 0.8959 &0.2012  & 25.01 (+0.14)    & 0.7788   &0.3204    &28.27 (+0.95) &0.8270 &0.3100               \\ \hline
CSNLN~\cite{mei}       & \textcolor{blue}{32.93 (+0.55)}              & \textcolor{blue}{0.9275}  &\textcolor{blue}{0.1357}        & 33.55 (+1.16)    & 0.9091 &0.1338 & 24.09 (+1.06)    & 0.7202 &0.3425 &28.79 (+0.43) &0.8275 &\textcolor{blue}{0.2343} \\ \hline
\end{tabular}
\caption{Quantitative comparison on (a) Parkour dataset, (b) Vimeo90K~\cite{vimeo90k} dataset and (c) Vid4 dataset. The metrics used are PSNR, SSIM and LPIPS. Larger numbers indicate better results for PSNR and SSIM, smaller numbers indicate better results for LPIPS. We also note the PSNR gain of our method comparing to other methods; a positive gain means that our method performs better than the corresponding method.}
\label{tab:quant}
\vspace{-15pt}
\end{table*}

\begin{table}[htb]
\scriptsize
\begin{tabular}{p{0.035\textwidth}<{\centering}|p{0.035\textwidth}<{\centering}|p{0.035\textwidth}<{\centering}|p{0.035\textwidth}<{\centering}|p{0.035\textwidth}<{\centering}|p{0.035\textwidth}<{\centering}|p{0.035\textwidth}<{\centering}|p{0.035\textwidth}<{\centering}}\label{tab:lpips}
            & PSNR & SSIM & LPIPS   &       & PSNR & SSIM & LPIPS   \\ \hline
\textbf{MANA}& \textbf{\textcolor{red}{38.62}} & \textbf{\textcolor{red}{0.9606}} & \textbf{\textcolor{blue}{0.1586}}& TGA &38.26                   & \textcolor{blue}{0.9588} & \textcolor{red}{0.1570}\\ \hline
EDVR & \textcolor{blue}{38.33} & 0.9544& 0.1641 & PFNL  &35.90                   & 0.9449  &0.1985 \\ \hline
TOFlow &36.55 & 0.9471 &0.1902 &CSNLN  & 37.79 & 0.9523  &0.1780 \\ \hline
\end{tabular}
\caption{Quantitative comparison on top 6\% large motion videos in Vimeo90K~\cite{vimeo90k} dataset.}
\label{tab:top5}
\vspace{-15pt}
\end{table}

\subsection{Datasets and Metrics}\label{subsec:dataset}
As discussed in Sec.~\ref{sec:intro}, the cross-frame non-local attention in our method enables VSR without frame alignment.
To validate the robustness of our method to large motion videos, we randomly collect 14 parkour video clips from the Internet.
Parkour is a form of extreme sport focusing on passing obstacles in a complex environment by running, climbing, and jumping.
Usually taken using egocentric wearable cameras, parkour videos are typical examples in the real-world where large camera motions are everywhere.
Example video stills from the \textit{Parkour dataset} are shown in Fig.~\ref{fig:dataset}.
We further evaluate our method on regular small motion videos using Vimeo90K~\cite{vimeo90k} test set and Vid4~\cite{frvsr}.
For all the test sets, we use the average PSNR and SSIM~\cite{ssim} on the RGB channels to quantitatively evaluate the performance of the methods.
In addition, we apply LPIPS~\cite{lpips} to evaluate the perceptual similarity between the super-resolved frames and the ground truth high-resolution frame.
Since the performance can be different across computation platforms and the quantitative metric calculation might be different in these works, we re-run their code and calculate the metrics in the same way on the same computer.

\subsection{Visual Comparisons}\label{subsec:visual}
The visual comparison of the examples from the Parkour dataset, Vimeo90K~\cite{vimeo90k} dataset and Vid4~\cite{frvsr} is shown in Fig.~\ref{fig:visual}.
To make the discussion concise, we label the ID at the bottom left of each video.
We also added arrows pointing at the regions we will be discussing.

Example (a), (b), (c), (d) and (e) are selected from the Parkour dataset.
These examples contain large motion and are challenging to existing VSR methods.
Our method can reconstruct repetitive patterns like Example (a) and (b), while explicit frame alignment methods TOFlow~\cite{vimeo90k} and TGA~\cite{tga} fail due to the inaccurate frame alignment.
EDVR~\cite{edvr} result is more blurry than our result in example (a) and (b), and \textit{the blurry issue is more visible when viewed in dynamics as we will show in the supplementary video}.
This indicates that the deformable convolution alignment cannot handle the alignment with large frame displacement.
The VSR and SISR methods PFNL~\cite{pfnl} and CSNLN~\cite{mei} using non-local attention also suffer from the blurry issue, potentially due to the non-local attention performance degradation problem discussed in Sec.~\ref{subsec:nonlocal}.

Example (c) focuses on general details of objects.
The frame-aligning VSR methods introduce ghosting artifacts (EDVR~\cite{edvr}) and deformation (TOFlow~\cite{vimeo90k} and TGA~\cite{tga}) due to the inaccurate alignment.
PFNL~\cite{pfnl} and CSNLN~\cite{mei} results have less detail than ours, indicating that our one-hot non-local attention improves the quality of regular non-local attention.
Example (d) focuses on human face shape and details.
As shown in the bicubic result, the original facial details are completely lost due to the down-sampling.
Our method reconstructs visually pleasing details of human faces thanks to the memory-augmented module, while the comparison methods introduce blur (EDVR~\cite{edvr}, TOFlow~\cite{vimeo90k} and PFNL~\cite{pfnl}) or reconstruct shapes that do not look like a human (TGA~\cite{tga} and CSNLN~\cite{mei}).

Example (e) contains thin structures.
Similar to examples (a) and (b), failure in frame alignment has negatively affected the VSR methods.
In this case, the performance of VSR methods EDVR~\cite{edvr}, TOFlow~\cite{vimeo90k} and TGA~\cite{tga} are even worse than the SISR method CSNLN~\cite{mei}.
Our method with one-hot non-local attention can achieve a comparable result to CSNLN~\cite{mei} in this example since our network does not require frame alignment.

As we will discuss in Sec.~\ref{subsec:quant}, the overall average quantitative metric score of our method is slightly inferior to that of EDVR~\cite{edvr} and TGA~\cite{tga} in the Vimeo90K~\cite{vimeo90k} and Vid4 dataset~\cite{frvsr} which are relatively easy for frame aligning VSR methods.
However, a larger deviation to the ground truth does not always indicate worse performance.
Example (f) and (g) are selected from the Vimeo90K~\cite{vimeo90k} test set.
Our method tends to produce visually sharper results than EDVR~\cite{edvr} and TGA~\cite{tga}, which is often more preferred in the video super-resolution task.
Example (h) is a widely used example in Vid4~\cite{frvsr}.
In this example, EDVR~\cite{edvr} and TGA~\cite{tga} generate the best results, but our result is comparable to their results.

To further evaluate the robustness of our method in real-world scenarios, we provide additional results in the supplementary material.
These videos are arbitrarily selected from different types of videos, covering animation, movies, and vlogs.
We encourage the readers to read the supplementary PDF for a more complete visual comparison.

\begin{table*}[t]
\footnotesize
\centering
\begin{tabular}{p{0.08\textwidth}||p{0.048\textwidth}<{\centering}|p{0.048\textwidth}<{\centering}|p{0.048\textwidth}<{\centering}||p{0.048\textwidth}<{\centering}|p{0.048\textwidth}<{\centering}|p{0.048\textwidth}<{\centering}||p{0.048\textwidth}<{\centering}|p{0.048\textwidth}<{\centering}|p{0.048\textwidth}<{\centering}||p{0.048\textwidth}<{\centering}|p{0.048\textwidth}<{\centering}|p{0.048\textwidth}<{\centering}}
            & \multicolumn{3}{c||}{Parkour Dataset} & \multicolumn{3}{c||}{Vimeo90K Dataset~\cite{vimeo90k}}    & \multicolumn{3}{c||}{Vid4 Dataset~\cite{frvsr}}  &  \multicolumn{3}{c}{SPMC Dataset~\cite{spmc}}     \\ \hline
            & PSNR (dB)          & SSIM    &LPIPS        & PSNR (dB) & SSIM &LPIPS  & PSNR (dB) & SSIM &LPIPS & PSNR (dB) & SSIM &LPIPS                        \\ \hline
No\_Mem     & 33.31              & 0.9343  &\textcolor{red}{0.1196}        & 34.47    & 0.9232 & 0.1696 & 25.09     & 0.7742  &0.3018 &29.03 &0.8389 &0.2256                    \\ \hline
$N=128$  & 33.44              & 0.9350     &0.1254     & 34.65       & 0.9254 &0.1886  & 25.13     & 0.7787  &0.2859 &29.24 &0.8449 & 0.2148                   \\ \hline
$\bm{N=256}$        & \textbf{\textcolor{red}{33.48}}              & \textbf{\textcolor{red}{0.9356}}    &\textbf{0.1241}      & \textbf{\textcolor{red}{34.71}}               & \textbf{\textcolor{red}{0.9261}} &\textbf{\textcolor{red}{0.1644}}  & \textbf{\textcolor{blue}{25.15}}     & \textbf{0.7796} &\textbf{\textcolor{red}{0.2744}} &\textbf{29.22} &\textbf{\textcolor{red}{0.8458}} &\textbf{\textcolor{red}{0.2119}}                     \\ \hline
$N=512$      & \textcolor{blue}{33.47}              & \textcolor{blue}{0.9354}   &0.1245       & \textcolor{blue}{34.71}               & \textcolor{blue}{0.9261} &\textcolor{blue}{0.1647}  & 25.14     & \textcolor{red}{0.7797}  &\textcolor{blue}{0.2840} &\textcolor{blue}{29.25} &0.8449 &\textcolor{blue}{0.2129}                    \\ \hline
$N=1024$         & 33.47 & 0.9354  &\textcolor{blue}{0.1238}  & 34.68   & 0.9257 &0.1649   & \textcolor{red}{25.15}     & \textcolor{blue}{0.7797}   &0.2852 &\textcolor{red}{29.26} &\textcolor{blue}{0.8452} &0.2142                   \\ \hline
\end{tabular}
\caption{Ablation study on the memory size in the memory-augmented attention module. The $N=256$ is selected for the experiments shown in Sec.~\ref{subsec:visual} and Sec.~\ref{subsec:quant}}
\label{tab:ablation}
%\vspace{-10pt}
\end{table*}

\subsection{Quantitative Comparisons}\label{subsec:quant}
In Table~\ref{tab:quant}(a), we compare the average PSNR and SSIM~\cite{ssim} of our method with the comparison methods on the Parkour dataset.
In this table, larger PSNR and SSIM and smaller LPIPS loss indicate better results.
We mark the best result in red and the second best result in blue.

The videos in the Parkour dataset have extremely large motions, making the accurate alignment of the frames difficult.
Among the comparison methods, TOFlow~\cite{vimeo90k} explicitly estimates the optical flow for warping neighbor frames; TGA~\cite{tga} uses homography to align neighbor frames; EDVR~\cite{edvr} implicitly align frames using learned kernel offset for deformable convolution.
The performances of these methods are even inferior to that of the SISR method CSNLN~\cite{mei} in the Parkour dataset, since fusing misaligned frames often cause blurry or ghosting artifacts in the result.
This indicates that the large motions have a negative effect on the performance of traditional VSR methods.

The comparison method PFNL~\cite{pfnl} uses pair-wise non-local attention on all the pixels in the entire segment.
As discussed in Sec.~\ref{subsec:nonlocal}, the traditional non-local attention is difficult to train due to the large GPU memory consumption and the performance degradation with a large number of pixels.
Although the performance of their method is better than frame alignment methods EDVR~\cite{edvr} and TGA~\cite{tga}, using traditional non-local attention on all video segments is worse than applying it only on a single frame like CSNLN~\cite{mei}.
Our cross-frame non-local attention mechanism significantly improves the performance of non-local attention by introducing the one-hot attention in the video super-resolution task.

In addition, we provide the quantitative comparison on the Vimeo90K dataset~\cite{vimeo90k}, the Vid4 dataset~\cite{frvsr} and the SPMC dataset~\cite{spmc} in Table~\ref{tab:quant}(b), (c) and (d) respectively.
The metrics and color labels are the same as Table~\ref{tab:quant}(a).
For these general small motion videos datasets, we also achieve better results than the explicit optical flow alignment VSR method TOFlow~\cite{vimeo90k} and the other non-local attention super-resolution methods PFNL~\cite{pfnl} and CSNLN~\cite{mei}.
Note that the single image non-local attention method CSNLN~\cite{mei} also outperforms video non-local attention method PFNL~\cite{pfnl} in the Vimeo90K dataset.
The PSNR and SSIM value of our method is slightly inferior to that of EDVR~\cite{edvr} and TGA~\cite{tga} in the Vimeo90K and Vid4 datasets.
EDVR~\cite{edvr} has a 0.97dB and 0.64dB PSNR gain to our method in Vimeo90K and Vid4 respectively.
TGA~\cite{tga} has a 0.32dB and 0.21dB PSNR gain to our method in Vimeo90K and Vid4 respectively.
The perceptual quality measured by the LPIPS~\cite{lpips} are similar among EDVR~\cite{edvr}, TGA~\cite{tga} and our method, with around 0.02 differences.
However, for the large motion examples in the Parkour dataset, our method has a larger PSNR gain (1.87dB and 2.34dB) and LPIPS gain (0.0659 and 0.0983) in the performance comparing to EDVR~\cite{edvr} and TGA~\cite{tga}.
Moreover, in Table.~\ref{tab:top5}, we computed the optical flow for videos in the Vimeo90K~\cite{vimeo90k} test set and ranked them based on the average flow magnitude.
It can be observed that even though EDVR~\cite{edvr} and TGA~\cite{tga} is slightly superior on average, our method is actually better on the large motion videos in Vimeo90K~\cite{vimeo90k}.

We also note that EDVR~\cite{edvr} is biased towards Vimeo90K~\cite{vimeo90k}, given the significant performance drop in the other small motion video dataset SPMC~\cite{spmc}. 
This implies that our method is more robust than the comparison methods.
We also provide the quantitative evaluation on the real-world videos in the supplementary material, which also supports the superior of our method in robustness.

\subsection{Ablation Study}\label{subsec:ablation}
In Table~\ref{tab:ablation}, we quantitatively compare the performance of different configurations in our network.
Specifically, we set the memory size $N$ of the memory-augmented attention module to 128, 256, 512 and 1024.
To verify the effectiveness of the memory-augmented attention module, we also experimented with the network with cross-frame non-local attention module only (labeled as \textit{No\_Mem} in Table~\ref{tab:ablation}).
Among these configurations, $N=256$ achieves the best result and is selected in the comparisons in Sec.~\ref{subsec:visual} and Sec.~\ref{subsec:quant}.
Using smaller memory ($N=128$) results in slight performance degradation.
The benefits saturate when using a larger memory ($N=512$ and $N=1024$), implying that the local details of low-resolution frames can be well represented in low-dimensional space.
The performance of our network degrades without the memory-augmented attention module.
However, solely using the cross-frame non-local attention module, our network outperforms comparison methods in the Parkour dataset and achieves comparable performance in the Vimeo90K dataset.

%% file: Conclusion.tex
\section{Conclusion}
We present a network for video super-resolution that is robust to large motion videos.
Unlike typical video super-resolution works, our network is able to super-resolve videos without aligning neighbor frames through a novel one-hot cross-frame non-local attention mechanism.
Thanks to the memory-augmented attention module, our method can also utilize information beyond the video that is being super-resolved by memorizing details of other videos during the training phase.
Our method achieves significantly better result in the large motion videos compared to the state-of-the-art video super-resolution methods.
The performance of our method is slightly inferior in the videos that are relatively easy for frame aligning video super-resolution methods.
We believe our method can be further improved by introducing pyramid structure into the cross-frame non-local attention to increase the perception field or extend the memory bank from 2D to higher dimension, but these ideas are left for future work.